\documentclass[letterpaper, 10 pt, conference]{ieeeconf}  

\IEEEoverridecommandlockouts                     		 

\usepackage{graphics} 
\usepackage{epsfig} 
\usepackage{mathptmx} 
\usepackage{times} 
\usepackage{amsmath} 

\usepackage{siunitx}
\usepackage{multicol} 
\usepackage{multirow}
\usepackage{cite}
\usepackage{amsmath,amssymb,amsfonts}
\usepackage{algorithmic}
\usepackage{graphicx}
\usepackage{textcomp}
\usepackage{xcolor}
\usepackage{amssymb}  
\usepackage{multirow}

\usepackage{url}
\usepackage{balance}
\title{\LARGE \bf
    CAMAL: Context-Aware Multi-layer 
    Attention framework for Lightweight Environment Invariant Visual Place Recognition
}

\author{Ahmad Khaliq$^{1}$, Shoaib Ehsan$^{1}$, Michael Milford$^{2}$ and Klaus McDonald-Maier$^{1}$
    \thanks{This work is supported by the UK EPSRC through grants EP/R02572X/1 and EP/P017487/1. The work is also supported by the RICE project funded by the NCNR Flexible Partnership Fund.}
    \thanks{$^{1}$Authors are with the Embedded and Intelligent System Lab in the School of Computer Science and Electronic Engineering, University of Essex, Colchester, United Kingdom
   	 {\tt\small \{ahmad.khaliq,sehsan,kdm\}@essex.ac.uk},{\tt\small\ ahmedest61@hotmail.com}.}%
    \thanks{$^{2}$Michael Milford is with the Australian Centre for Robotic Vision and School of Electrical Engineering and Computer Science, Queensland University of Technology, Brisbane, Australia
   	 {\tt\small michael.milford@qut.edu.au}.}%
}

\begin{document}
  
    \maketitle
    \thispagestyle{empty}
    \pagestyle{empty}

    \begin{abstract}
   
   
   In the last few years, Deep Convolutional Neural Networks (D-CNNs) have shown state-of-the-art (SOTA) performance for Visual Place Recognition (VPR), a pivotal component of long-term intelligent robotic vision (vision-aware localization and navigation systems). The prestigious generalization power of D-CNNs gained upon training on large scale places datasets and learned persistent image regions which are found to be robust 
   for specific place recognition under changing conditions and camera viewpoints. However, against the computation and power intensive D-CNNs based VPR algorithms that are employed to determine the approximate location of resource-constrained mobile robots, lightweight VPR techniques are preferred. This paper presents a computation- and energy-efficient CAMAL framework that captures place-specific multi-layer convolutional attentions efficient for environment invariant-VPR. At $4x$ lesser power consumption, evaluating the proposed VPR framework on challenging benchmark place recognition datasets reveal better and comparable Area under Precision-Recall (AUC-PR) curves with approximately $4x$ improved image retrieval performance over the contemporary VPR methodologies. 
   	 
   	 
   	 
    \end{abstract}
    
\keywords 
Convolutional Neural Network, Context-based Regional Attentions, Robot Localization, Visual Place Recognition.
\endkeywords

    \section{INTRODUCTION}
    
    Using a visual sensor, the process to estimate a location within a well-known environment is termed as Localization and finding a previously-visited location within that map is Visual Place Recognition (VPR) \cite{lowry2016visual}. Localization against the topological map of the environment can be performed in parallel to create and update that map, a problem known as Simultaneous Localization and Mapping (SLAM) \cite{cadena2016past}. In SLAM systems, an approximate location or global localization can be achieved through loop-closure detection, also researched as VPR and prime focus of this paper. Data association for image retrieval-based VPR techniques require a compact feature representation; earlier handcraft-based techniques (SURF \cite{bay2006surf}, SIFT \cite{ke2004pca} and HOG \cite{dalal2005histograms}) coupled with feature encoding methodologies including Bag of Words (BoW) and Vector of Locally Aggregated Descriptors (VLAD) demonstrated superior recognition performances. However, they are unable to deal with simultaneous visual changes experienced in day-night/summer-winter transitions and camera viewpoint variation. Deep convolutional neural networks (D-CNNs) with rich spatial and change-invariant features from middle and late convolutional layers \cite{chen2014convolutional}\cite{sunderhauf2015performance}\cite{sunderhauf2015place} have rescued the task of VPR under simultaneous variations observed in scene appearance and camera viewpoint. 
    

     \begin{figure}[t]
		\centering
	    \includegraphics[width=0.8\linewidth, height=0.65\linewidth,keepaspectratio]{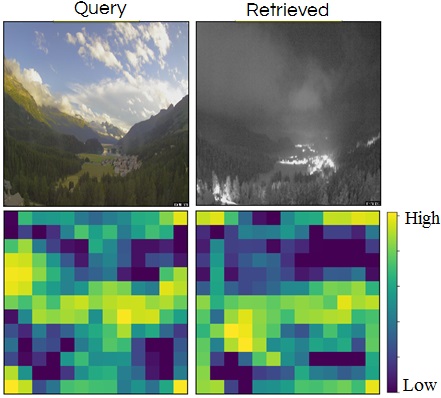}
    	\caption{
    	In this paper, our proposed CAMAL framework captures place-specific multi-convolutional attentions persistent under changing environment (illustrated with color heatmap) which primarily helps to recognize the place used for global localization of mobile robots.
    	The system takes a query as an input and returns a matched database image at $4x$ lesser power consumption \& quicker retrieval time over contemporary VPR algorithms. 
    	}
    	\label{figure:topMatchedImg}
    	\vspace{-5mm}
    \end{figure}

    Employing a CNN pre-trained on task dependent dataset and identifying meaningful image regions has been an area of ongoing research including image classification and retrieval problems \cite{wang2017residual}\cite{wang2017multi}. Likewise in VPR problems \cite{chen2014convolutional}\cite{sunderhauf2015performance}\cite{sunderhauf2015place}, employing off-the-shelf pretrained CNNs for finding cues based on vital image regions under changing environment has been a great interest in robotics and computer vision communities \cite{lowry2016visual}\cite{cadena2016past}. However, such pre-trained CNNs are different in nature from recognizing the places under seasonal, lighting and viewpoint changes where activations are non-uniformly distributed over the convolution layers as compared to tasks \cite{krizhevsky2012imagenet} where a single object occupies the whole image. Zhou \textit{et al.} in \cite{zhou2017places} trained a CNN model using a $10$ Million Places365 dataset for the task of scene recognition (place categorization). Place categorization problem resembles with place recognition given that environments are considered to determine the type of place from a stored scene database. However, an image under the same category of scene (kitchen) can represent a geographically different place, thus only categorizes the ``place scene" but not allocate the ``exact place". Specific place recognition is the key to perform loop-closure detection needed for mobile robotics to approximate their location under strong environmental variations \cite{torii201524} where human involvement is out of question.  

    Solving VPR problem in \cite{chen2017only}, Chen \textit{et al.} employed a deep neural network VGG-16 \cite{simonyan2014very} pre-trained on an object-centric ImageNet \cite{krizhevsky2012imagenet} dataset and used middle convolutional layers for feature extraction based on the identified regions from late convolutional layers. 
    Later, Chen \textit{et al.} in \cite{chen2018learning} fine-tuned the pre-trained object-centric VGG-16 \cite{simonyan2014very} on the Specific PlacEs Dataset (SPED) \cite{chen2017deep}. 
    A context-flexible block is integrated inside the fine-tuned deep feed forward neural network to learn place-specific regions of interest (ROIs). 
    Khaliq \textit{et al.} in \cite{khaliqholistic} have introduced a lightweight novel approach for extracting regions-based convolutional feature from a shallow CNN AlexNet365 pretrained on Places365 dataset \cite{zhou2017places}. They have used middle convolutional layer for regional feature extraction coupled with VLAD \cite{jegou2010aggregating} encoding. 
    The Region-VLAD \cite{khaliqholistic} VPR framework has shown a performance boost in Area under Precision-Recall curves (AUC-PR)
    on several viewpoint- and condition-variant datasets against the SOTA contemporary VPR algorithms including FAB-MAP \cite{cummins2008fab}, SeqSLAM \cite{milford2012seqslam}, R-MAC \cite{tolias2015particular}, Cross-Region-BoW \cite{chen2017only} and convolutional feature pooling methodologies like Sum-Pooling \cite{babenko2015aggregating}, Max-Pooling \cite{tolias2015particular} and Cross-Pooling  \cite{liu2017cross}. 
    
    Focusing on dynamic entities (pedestrians, vehicles) other than the static landmarks can be efficient in dealing with the scene recognition \cite{zhou2017places} but it can instigate deceptive information in recognizing specific places due to perceptual aliasing \cite{milford2012seqslam} which leads to incorrect places association. Despite the better recognition performances of \cite{chen2017only}\cite{khaliqholistic} VPR algorithms, both suffered with the inclusion of dynamic objects in the regions-based CNN representation. 
    To address this problem, we extend the idea of \cite{khaliqholistic} to multi-layer regional approach at $3x$ reduction in feature encoding time and integrated on the pretrained place recognition-centric HybridNet \cite{chen2017deep} CNN model. 
    The proposed VPR framework captures powerful and rich semantic multi-convolution regional attentions where the attentions' areas vary with the place context. Employing D-CNNs, the authors in \cite{chen2017deep}\cite{chu2017multi}\cite{jin2017learned}\cite{chen2018learning} also attempt to learn fused multi-level regional features for environment invariant-VPR. 
    However, improving specific place recognition performance with D-CNNs add computational and power constraints for battery-powered robotic platforms where response time is vital \cite{cadena2016past}. 
    Figure \ref{figure:topMatchedImg} illustrates the place-specific multi-layer convolutional attentions identified by our proposed image retrieval-based VPR framework on the query and retrieved database place. 
    Our main contributions of the work are as follow:
    \begin{enumerate}
    \item The proposed lightweight VPR framework is integrated with HybridNet CNN model and taking precedence of CNN's fine-tuning on condition-variant SPED, the technique captures place-specific multi-convolutional attentions which remain persistent under strong visual changes in the presence of confusing instances.  
    \item A range of experiments on the challenging place recognition datasets exhibiting strong environmental variations confirm better \& comparable Area under PR-curves at $4x$ faster image retrieval performance and $4x$ lower power consumption over SOTA VPR algorithms. 
    
\end{enumerate}
   The rest of the paper is organized as follows. Section II provides a literature review of both handcrafted and CNN-based VPR paradigms. In section III, we describe the proposed framework in detail. Section IV presents the benchmark datasets, evaluation metrics, experimental setup, performance evaluation, detailed analysis and results. Section V ends with the conclusion and future work.

    \section{Literature Review}
    
    In a VPR system, image processing is the first module involved in identifying and extracting distinguishing features. The early approaches consisted of human-made feature detection techniques \cite{bay2006surf}\cite{ke2004pca}\cite{dalal2005histograms}, classified into local or global representations. Scale Invariant Feature Transform (SIFT) \cite{ke2004pca}, a local feature detector that extracts and describes the keypoints using difference-of-gaussian and histogram-of-oriented-gradients. Other approaches include HOG, SURF, FAST \cite{rosten2006machine}, GIST \cite{oliva2006building}, FABMAP \cite{cummins2008fab} and SEQSLAM \cite{milford2012seqslam}. FABMAP, a combination of SURF features with Bag-of-Words (BoW) \cite{sivic2003video} encoding scheme demonstrated robustness against viewpoint change. SEQSLAM is another appearance-invariant VPR technique that subtracts patch-normalized frames captured in a sequence followed up with a confusion matrix for best match retrieval.
    
    The recent boom of deep learning in various computer vision \cite{wang2017residual}\cite{wang2017multi} and robotic \cite{lowry2016visual}\cite{cadena2016past} platforms inspires and opens up the research gate for the VPR community. Chen \textit{et al.} in \cite{chen2014convolutional} for the first time used CNN-based features for the VPR problem. Work in \cite{sunderhauf2015performance}\cite{sunderhauf2015place} provided a detailed analysis of middle and late convolutional layers' robustness for specific place recognition. Authors in \cite{panphattarasap2016visual}\cite{sunderhauf2015place} combined CNNs with external landmark-based approaches. All aforementioned techniques generally employed CNN models pre-trained on tasks other than place recognition. To compensate this research gap, Chen \textit{et al.} in \cite{chen2017deep} have introduced and evaluated the performance of two place recognition-centric CNNs for VPR; AMOSNet and HybridNet, pre-trained and fine-tuned the object-centric CaffeNet \cite{krizhevsky2012imagenet} on $2.5$ Million SPED \cite{chen2017deep}. The results claimed that Spatial Pyramid Pooling (SPP) on late convolutional layers of fine-tuned HybridNet has shown a performance boost over AMOSNet, CaffeNet and scene-centric PlaceNet evaluated on several challenging place-recognition datasets. Arandjelovic \textit{et al.} in \cite{arandjelovic2016netvlad} added a VLAD layer in the CNN architecture, named NetVLAD and trained the model end to end on 
    Pittsburgh dataset. 
    
    Training and fine-tuning the CNNs on large scale task-dependent datasets induce change-invariance in the convolutional layers. However, employing pretrained CNNs directly cannot boost up the performance. In literature, various convolutional feature pooling techniques including Sum- \cite{babenko2015aggregating}, Max- \cite{tolias2015particular}, Spatial Max- \cite{chen2017deep} and Cross-Pooling \cite{liu2017cross} are proposed and shown performance boost for image retrieval \cite{wang2017multi} and classification \cite{wang2017residual} but found not to suitable for place recognition \cite{chen2017only}. One of the focus of this work is to employ a place recognition-centric CNN and pick place-specific convolutional attentions robust under changing environment needed for global localization of mobile and aerial robots.  
    
    Zaffar \textit{et al.} in \cite{zaffar2018memorablemaps} introduced a cognition-inspired agnostic framework that evaluates the goodness of the place used for the task of VPR. Recent VPR research work in \cite{chen2017only}\cite{chen2018learning}\cite{facil2019condition}\cite{8972582} employed regions-based feature description either employing handcraft-based techniques or D-CNNs. 
    Cross-Region-BoW \cite{chen2017only} used a similar idea of cross-convolution \cite{liu2017cross} over late convolutional layers of deep VGG-16 and employed $200$ ROIs coupled with $10k$ BoW dictionary. 
    RMAC \cite{tolias2015particular} also employed a regional approach based on maximum activations of convolution. Siagian \textit{et al.} in \cite{siagian2009biologically} employed an attention-based approach for mobile robots. Authors in \cite{chen2016attention}\cite{wang2017residual} further demonstrated that attention-based features can play an important role in improving vision-based robotics tasks. However, such attention capturing techniques require manually defined regional masks. Zaffar \textit{et al.} in \cite{8972582} have proposed a lightweight and training-free VPR framework, namely CoHOG that captures covolutional regional features based on HOG descriptor. The CoHOG technique is evaluated on camera viewpoint- and condition-variant place recognition datasets and achieved SOTA Performance-per-Compute-Unit (PCU) at $20x$ lower feature encoding time. 
    
    Motivated from the work of \cite{kim2017learned} which overcomes the difficulty of manually employing a fixed regional attention mask over deep CNNs, Chen \textit{et al.} in \cite{chen2018learning} have integrated a context-flexible attention block in a deep object-centric VGG-16 and fine-tuned it on condition-variant SPED. The proposed system was trained end-to-end specifically for VPR problem. 
    However, all the aforementioned contemporary VPR techniques used deep VGG-16 model and employed late convolutional layers for feature extraction which means high power consumption and computational resources are required for robotic platforms at execution time. Khaliq \textit{et al.} in \cite{khaliqholistic} bridged this research gap with a lightweight but manual regional approach which can be incorporated within any CNN model. 
    At low computation and resource utilization, the Region-VLAD approach in \cite{khaliqholistic} employed middle convolutional layers of AlexNet365 and shown SOTA performance over \cite{chen2017only} in terms of AUC-PR curves. 
    
\begin{figure}[t]
		\centering
	    \includegraphics[width=1\linewidth, height=1\linewidth,keepaspectratio]{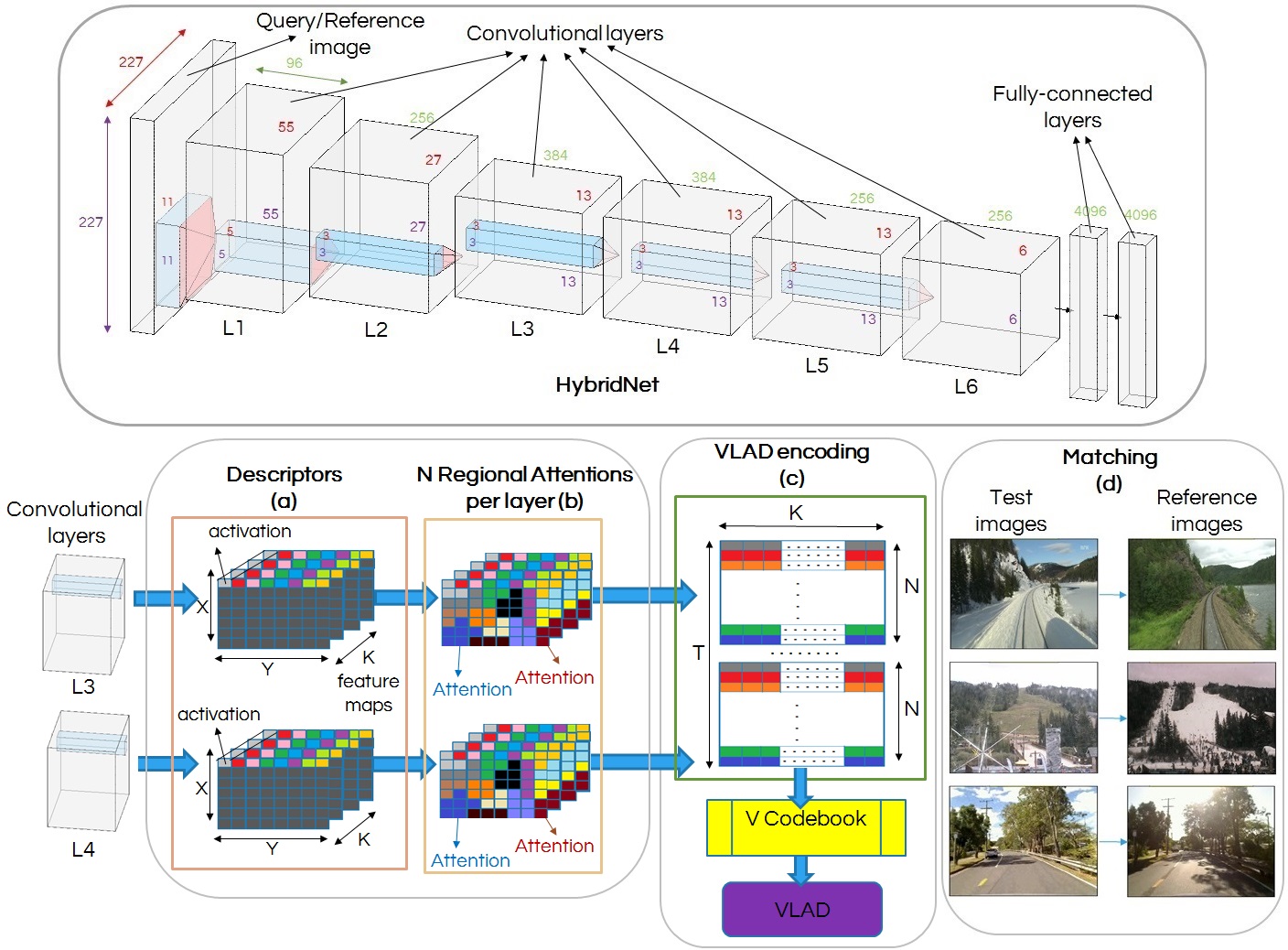}
		\caption{Our proposed CAMAL VPR framework is employed on HybridNet CNN model (its layers' architecture is shown at the top). Given an image of the place, we process $L3$ and $L4$ layers only (shown at the bottom). Convolutional activations are stacked to retrieve the spatial local descriptors (a). $N=300$ salient regions in (b) are identified per layer and using a $V=128$ regional codebook pretrained on a separate dataset, the multi-layer $T$ regional attentions in (c) are used for VLAD encoding and matching (d). 
		} 
		\label{figure:proposedBlock}
		\vspace{-4mm}
	\end{figure}

    Despite the supreme matching performance of \cite{khaliqholistic}, it sometimes include dynamic instances in the captured CNN-based regional features which are deleterious for recognizing specific places under changing environment. 
    Authors in \cite{chen2016attention}\cite{chen2017deep}\cite{chu2017multi} demonstrated that attentions based on multiple convolutional layers are robust under changing environment (and perceptual aliasing) in the presence of dynamic instances. 
    Taking inspiration from \cite{khaliqholistic}\cite{chen2016attention}, we have optimized the regional approach of \cite{khaliqholistic} and integrated at multiple convolutional layers of shallow CNN SPED-centric HybridNet shown in Fig. \ref{figure:proposedBlock}. The upper part exhibits the architecture of HybridNet and the lower part illustrates the CAMAL framework. 
    We have shown in the experiments that our proposed framework captured meaningful image regions invariant to strong conditional and moderate viewpoint variations. At $4x$ lesser power utilization, evaluation on several benchmark place-recognition datasets have shown better and comparable AUC-PR curves performance with $4x$ faster image retrieval time over various VPR techniques \cite{tolias2015particular}\cite{arandjelovic2016netvlad}\cite{chen2017only}\cite{chen2018learning}\cite{khaliqholistic}.  
    \vspace{-1mm}
    \section{Proposed Technique}
    This section describes the proposed framework in more detail. To subdivide an image into spatial convolutional regions, we first discuss the retrieval of local descriptors from the convolutional feature maps. We then demonstrate our approach of finding regional attentions from multiple convolutional layers of HybridNet CNN and discussion on obtaining their VLAD representation for image matching. 
    
    \vspace{-1mm}

    \subsection{Local Descriptors of Convolutional Activations}
    \label{featuresStacking}
    
    In a neural network, $ X \times Y \times K $ is the dimension of $ 3D $ convolutional layer tensor $ M $, where $ X $ and $ Y $ represent the width and height of each channel and $ K $ is the number of channels, also termed as feature maps. In layman terms, each feature map $k = \{ 1,2,....,K\}$ corresponds to some filter being convolve on the input image $I$. At a certain spatial location $(i,j)$, we stack down the activations of $K$ feature maps, and each spatially stacked $1-D$ activations vector is termed as a local descriptor, (see Fig. \ref{figure:proposedBlock}(a)). In eq (\ref{eq:1}), $D^L$ denotes the $K$ dimensional local descriptors at $L^{th}$ convolutional layer. 	 
    \vspace{-1mm}
	\begin{equation}D^L =\{d_l  \in  M^K\ \ \forall  \ l \in \{(i,j) \ | \ i=1, ...,X; j=1, ..., Y\}\}\vspace{-1mm}\label{eq:1}\end{equation}

    \subsection{Identification of Place-Specific Regional Attentions}
    \label{ROI}
    
    
    Within the convolutional layer of the CNN model, certain spatial regions of feature maps do have more intensity mimicking the presence of certain visual patterns in the image. For example, giving an image of an urban/rural road scene as an input to a CNN, one certain convolutional feature map might be focusing on the vehicles while others can find buildings as an important visual clue. In the context of specific place recognition, focusing on time-varying objects such as pedestrians and vehicles can degrade the overall matching performance. Therefore, salient regions corresponding to contextual meaningful attentions including road signals, buildings can be useful to recognize a specific place under simultaneous variations in condition and viewpoint.
    
    To identify and capture CNN-based place-specific distinguishing attentions, we chose to employ 8-layered CNN SPED-centric HybridNet \cite{chen2017deep}, as illustrated in Fig. \ref{figure:proposedBlock}. It's the fine-tuned version of the object-centric CaffeNet on a large scale $2.5$ Million Specific PlacEs Dataset (SPED). SPED contains thousands of geographically different places (labels), and each label contains hundreds of images of the same place captured from surveillance cameras under severe environmental conditions. Classification-based fine-tuning of the CNN on specific places learned contextual condition-invariant features and with our convolutional region-based framework, we captured place-specific regional attentions that are powerful and robust for environment invariant-VPR. 
  
  	Particularly, we pass down an image into the HybridNet and individually process the feature maps of $L3$ and $L4$ convolutional layers only. Non-zero spatially connected activations per feature map are grouped such that if two or more activations roughly have similar responses then they couple to represent a $G_h$ region, $ \forall \ h \in \{1,...,H\} $ where $H$ is the total number of identified regional attentions from $K$ feature maps at $L^{ th}$ convolution layer. Similar to \cite{khaliqholistic}, energies of all the identified attentions are calculated by averaging over all the $a_h$ activations lying under each $G_h$ attention. In eq (\ref{eq:2}), $a^f_h$ represents the $f^{th}$ activation lying under $G_h$ region and $E^L$ denotes the energies of all the regions. In eq (\ref{eq:3}), with sorted $E^L$ energies, $R^L$ represents the top $N$ energetic novel context-aware attentions at $L^{th}$ convolutional layer. 

    \vspace{-3mm}
    


    \begin{equation}\vspace{-3mm} E^L = \{ \frac{1}{|G_h|} \ \sum\limits_f a^f_h,\ \forall \ a^f_h \in G_h\} \label{eq:2}\end{equation}

    \begin{equation}\vspace{-1mm}R^L = \{G_t \ \forall  \ t \in \{1,...,N\}\}\label{eq:3}\end{equation}

  To improve the recognition performance, it is important that the identified regions should be distinctive and persistent, i.e., environment-dependent regions should not be included otherwise it is highly likely that the system recognizes the scene correctly but the matched location is geographically different. Experimentation at $N=300$ confirms minimal dynamic instances since with the inclusion of more but less energetic regions, activations concentrated on dynamic objects do get included. Under the identified $q$ regional attention, $D^L_q$ denotes the underlying regional local descriptors, aggregated in (\ref{eq:4}) to retrieve a single $1\times K$ dimensional convolutional regional feature and $N \times K$ dimensional $f^L$ represents all the $N$ regional features. $F_I$ in (\ref{eq:5}) is $T \times K$ dimensional regional features captured from $L_3$ and $L_4$ convolutional layers, illustrated in Fig. \ref{figure:proposedBlock}(c).

    \vspace{-3mm}
    
    \begin{equation}f^L = \{\sum\limits_{q \in R^L_t} D^L_q \ \forall \ t \in \{1,...,N\}\}\label{eq:4}\end{equation}
    \vspace{-3mm}
    \begin{equation}F_I = f^L \ \forall  \ L \in \{L_3,L_4\}\vspace{-1mm}\label{eq:5}\end{equation}

\begin{figure}[h]
	\centering
	\includegraphics[width=1.0\linewidth, height=1\linewidth,keepaspectratio]{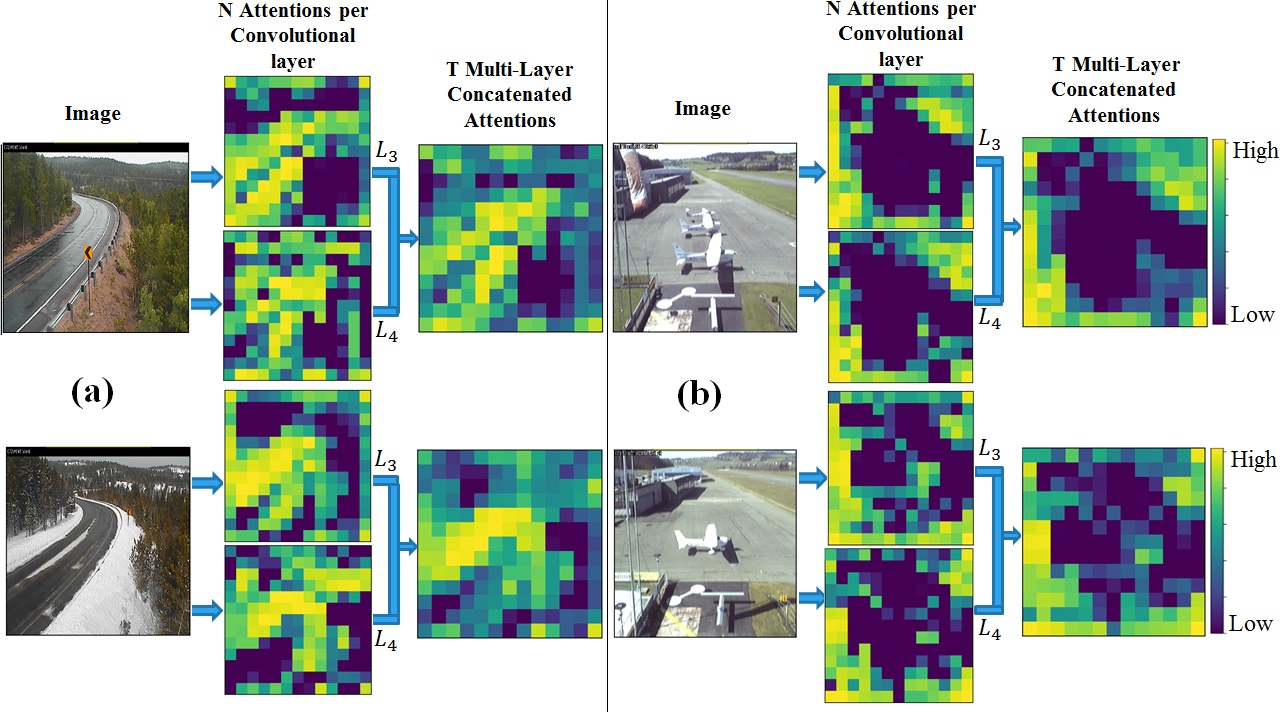}
	\caption{For place (a) and (b) captured twice and experienced strong conditional change in the presence of dynamic objects, $T$ multi-convolutional attentions are sorted based on their energies (using (\ref{eq:2})). The color heatmaps illustrate the identified place-specific single- \& multi-convolutional attentions captured by our proposed environment invariant-VPR framework. 
	}
	\label{figure:ROIs}
	\vspace{-4mm}
\end{figure}

	To further analysis and provide the insight into how the multi-layer regional attentions can be useful under external environmental variations, $F_I$ attentions are sorted based on their energies (eq (\ref{eq:2})) and their regional impact is highlighted in Fig. \ref{figure:ROIs}; place (a) and (b) are captured twice under different conditions. 
	A closer look reveals that the sorted multi-layer convolutional attentions reduce down the impact of fewer regions (identified by individual layer) focusing on dynamic instances such as plane and tree. 
	It suggest that employing multiple convolutional layers of HybridNet, our CAMAL framework filters and captures the low and high level regional features focusing on meaningful contextual place-specific regional attentions which remain persistent and robust under strong environment and mild viewpoint variations.  


    \subsection{Attention-based Codebook \& VLAD for Image Matching}
    \label{vocabVLAD}
    With smaller visual word vocabulary in tasks including image retrieval, recognition and object detection \cite{tolias2015particular}\cite{liu2017cross}, VLAD \cite{jegou2010aggregating} has shown SOTA performance. For attention-based dictionary, we have collected a separate dataset of $3000$ images which consist of $1125$ Query247 \cite{torii201524} images taken in day, evening and night times of $365$ places. The other images include  Garden Point dataset\cite{chen2017deep}, Eynsham dataset\cite{chen2014convolutional} and multiple condition- and viewpoint-variant rural and urban road traverses captured from \textit{Mapillary} \cite{sunderhauf2015place}\cite{chen2017only}. 
    Similar to \cite{khaliqholistic}, K-means is used to cluster $3000 \times T \times K$ dimensional multi-layer $F_I$ convolutional attentions into $K$-dimensional $V=128$ regions. For all the benchmark test and reference frames, their attentions are quantized to predict the dictionary clusters/labels. The $128\times K$ dimensional VLAD descriptor is obtained using the multi-layer $F_I$ attentions, predicted labels and attention-based pre-trained vocabulary.

    \section{Datasets, Comparison \& Evaluation Criteria, Implementation Setup, Performance Evaluation, Results and Analysis}
    This section discusses the benchmark datasets employed to determine the proposed VPR framework efficiency of recognizing specific places under strong environment and significant camera-viewpoint changes against the contemporary VPR methodologies. We first highlight the run-time implementation details followed up with the discussion on the performance metrics and evaluation. We also compare the multi-layer place-specific regional attentions identified by CAMAL and SOTA VPR techniques \cite{chen2018learning}\cite{khaliqholistic}.

    \vspace{-1mm}
    \subsection{Benchmark Place Recognition Datasets}
    We have employed three challenging place recognition datasets (see Table \ref{table:datasets}) to evaluate the VPR approaches. Two traverses along the same route taken at multiple times of the day/year are captured per dataset exhibiting scenarios experience by mobile robots in real world. The first traverse is used for testing and the second traverse is served as reference. The St.Lucia dataset \cite{chen2017deep} was captured in the suburban route at multiple day times with sufficient viewpoint- and condition-variation. The provided GPS annotation to build place and frame level correspondence is used for ground truth. 
    
    The SPEDTest \cite{chen2018learning} is the newly introduced dataset and contains diverse scenarios captured with surveillance cameras in multiple time of the year (for more information, please see \cite{chen2018learning}). There is a strong illumination change with mild viewpoint variance and for the ground truth, each test image resembles with three known reference images. The Synthesized Nordland dataset \cite{chen2018learning} is a mild viewpoint-variant version of original Synthesized Nordland \cite{Neubert2014SuperpixelbasedAC} with 75\% resemblance among the traverses. It's a train journey captured in winter and summer seasons, with frame and place level resemblance is used for ground truth.   
%
    \renewcommand{\arraystretch}{0.9}
    \renewcommand{\tabcolsep}{2.9pt}
    \begin{table}[h]
    	\vspace{-2.5mm}
    	\caption{Benchmark place recognition datasets}
    	\label{table:datasets}
   	 
    	\begin{tabular}{|c|c|c|c|c|c|}
      	  \hline
      	  \multirow{2}{*}{\textbf{Dataset}}                        		 & \multicolumn{2}{c|}{\textbf{Traverse}} & \multirow{2}{*}{\textbf{Environment}}             		 & \multicolumn{2}{c|}{\textbf{Variation}} \\ \cline{2-3} \cline{5-6}
      	  & \textbf{Test}   & \textbf{Reference}   &                                                   		 & \textbf{Viewpoint} & \textbf{Condition} \\ \hline
      	  St. Lucia                                              		 & 1249     		 & 1249          		 & Suburban                                             		 & Adequate    		 & Significant  		 \\ \hline
      	  \begin{tabular}[c]{@{}c@{}}SPEDTest \end{tabular} & 607    		 & 1821          		 & \begin{tabular}[c]{@{}c@{}} Diverse \end{tabular} & Moderate   	 &  Strong  		 \\ \hline
      	  \begin{tabular}[c]{@{}c@{}}Synthesized \\ Nordland \end{tabular} 		 & 1622    		 & 1622         		 & Train journey                                             		 & Moderate   	 & Very Strong   		 \\ \hline

    	\end{tabular}
    	\vspace{-2mm}
    \end{table}
    
    \vspace{-2mm}
    
    \subsection{Comparison VPR Frameworks}
    \label{section:methods}
    
    To make a fair comparison, we have reported the recognition performance of VPR approaches evaluated in \cite{chen2018learning}; Attentive Attention \cite{yu2017multi}, Cross-Pool, FABMAP, Fix-Context \cite{jin2017learned}, Context Flexible Attention, Places365 \cite{zhou2017places} and SEQSLAM. Particularly, for SOTA Attentive Attention approach and VPR-based Fix-Context framework, Chen \textit{et al.} in \cite{chen2018learning} have fine-tuned these models on SPED. For Cross-Pool \cite{liu2017cross}, the late convolutional layer is employed to generate a fixed attention mask, used as image feature. For handcraft-based VPR approaches; FABMAP and SEQSLAM, the authors have used their official implementations \cite{cummins2009highly}\cite{milford2012seqslam}. Places365 is a CNN model pre-trained on a \textit{2 Million} diverse dataset consisting of scenes for the task of scene recognition. The responses of late fully-connected layer are used as features. 
    
    Furthermore, other CNN-based VPR algorithms such as NetVLAD, RMAC, Cross-Region-BoW and Region-VLAD are also evaluated. For Region-VLAD, $N=200$ regions are employed from \textit{conv3} of AlexNet365 with $V=128$ clustered vocabulary for VLAD retrieval \cite{khaliqholistic}. All other approaches used VGG-16 pre-trained on object-centric ImageNet database. Their layers configuration are kept same as reported in \cite{zaffar2019levelling}; \textit{conv5\_2} is used for RMAC, with power- and l2-normalization on the regional features. For Cross-Region-BoW, \textit{conv5\_2} and \textit{conv5\_3} are employed with $10k$ BoW dictionary. For both the techniques, cosine matching is performed for filtering the mutual regions and their scores are summed and database image with highest score considered as matched place. Given an image, NetVLAD outputs a $1D$ feature descriptor and one-to-one cosine matching is performed followed up with scores summation. The reference image with highest score is considered as the current place.   
    \vspace{-1mm}
    \subsection{Evaluation Metrics}
    \label{section:eval_criteria}

     \subsubsection{Image Retrieval time \& Power Consumption}
     In autonomous robotic platforms where VPR is used for approximate/global localization purpose, total time required for image retrieval (or recognition) remains very crucial. 
     To report the $M_q$ query retrieval time, we calculate the feature encoding time ($M_f+M_e+M_v$) and feature matching time $M_m$ for the given query against $R$ reference images, mathematically represented in eq (\ref{eq:15}). $M_f$ represents the forward pass time for an image through a CNN model (if applicable), feature extraction time $M_e$, feature/VLAD encoding $M_v$ and $M_m$ denotes one-to-one feature/VLAD matching time.
     
     \vspace{-2mm}
	\begin{equation}M_q = M_f + M_e + M_v + M_m*R \label{eq:15}\vspace{-1mm}\end{equation}

	The other metric we have used to evaluate the VPR techniques is battery/power consumption, a vital component for resource-constrained mobile and aerial robots. Inspired by the recent work in \cite{zaffar2019state} where Zaffar \textit{et al.} have introduced and reported the power consumption of various VPR algorithms running on a common platform. We adopt the mathematical notation in (\ref{eq:16}) and report the total battery consumed in $miliampere-hours$ given that $Q$ number of queries are matched against $R$ number of reference images.  
	
	\vspace{-2mm}
	\begin{equation}mAh =  \frac {(U_e \times t_e + U_m \times t_m) \times Q}{v \times 60 \times 60}  \label{eq:16}\end{equation}
	
	$U_e$ and $t_e$ in (\ref{eq:16}) denote the CPU utilization and feature encoding time (milisec) per query image whereas $U_m$ and $t_m$ represent the CPU utilization and feature matching time (milisec) for each query against $R$ reference images and $v(=2.5)$ in denominator is the constant voltage used by the CPUs. Other parameters are kept same as reported in \cite{zaffar2019state}.
	
     \subsubsection{Area under Precision-Recall curves}
	For image retrieval based VPR systems, Area under Precision-Recall curves \cite{hanley1982meaning} is widely employed to evaluate the recognition performance. $Precision$ determines the proportion of correctly retrieved positive images. $Recall$ indicates the proportion of actual positive images correctly identified. 
	We used the python sklearn library to determine the AUC-PR curves.     
	
    \subsection{Implementation details}
    Deep learning techniques are computationally expensive which makes it indispensable to evaluate the run-time performance to realize its deployment in robotic VPR platforms. Evaluation of all the VPR techniques are performed on \textbf{Intel Xeon(R) Gold 6134 CPU @ 3.20GHz with 32 cores and 64GB RAM}. The proposed CAMAL VPR framework is implemented in Python $3.6.4$ (Caffe) and the system average run-time parameters over $3$ iterations with $3244$ images ($Q=1622$, $R=1622$) are recorded, see Table \ref{table:vprComparison}. We employed HybridNet CNN model, load (and run) its weights till middle $L3=conv3$ and $L4=conv4$ layers to capture multi-convolutional context-aware regional features. 
    \subsection{Performance Evaluation}
    \subsubsection{\textbf{Image Retrieval time \& Power Consumption}}
    Passing an image into the HybridNet, the forward pass takes an average $M_f = 13.85 \ ms$ for CAMAL. $N=300$ attentions per layer and $V = 128$ clustered vocabulary for VLAD encoding are used. Extraction of $T(=N*2)$ multi-layer attentions per image takes around $ M_e = 110 \ ms$ with VLAD encoding and one-to-one matching consume $ M_v = 2.68 \ ms$ and $ M_m = 0.07 \ ms$ \cite{khaliqholistic}. The overall feature encoding takes $M_f + M_e + M_v = t_e = 126.53 \ ms$ as reported in Table \ref{table:vprComparison}. 
    
\begin{figure}[h]
	\centering
	\vspace{-2mm}
	\includegraphics[width=1\linewidth, height=1.2\linewidth,keepaspectratio]{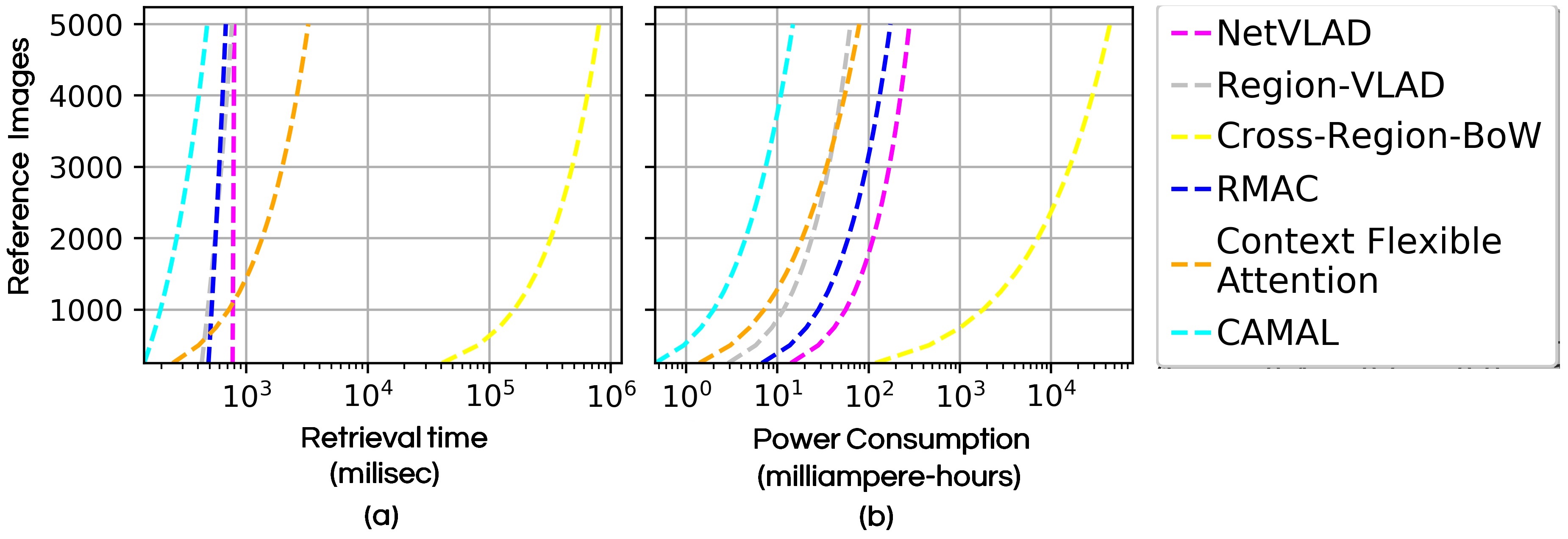}
	\caption{For all the contemporary VPR techniques given $R=\{250,500,...,5000\}$ reference images (a): Image retrieval time $M_q$ for a single query (b): Power consumption $mAh$ for matching $Q=1622$ queries.    
	}
	\label{figure:imgRet}
	\vspace{-3mm}
	
\end{figure}

    Evaluation with $R=\{250,500,750,...,5000\}$ reference images for $M_q$ retrieval time per query and $mAh$ power consumption for matching $Q=1622$ queries, we plot their logarithmic scale graphs for all the contemporary VPR techniques shown in Fig. \ref{figure:imgRet}. Other parameters for power calculation including $U_e$ and $U_m$ are measured as reported in \cite{zaffar2019state}. In Table \ref{table:vprComparison}, we have used $R=1622$ to report $M_q$ retrieval time per query and $Q=1622$ for $mAh$ power consumption, comes around $240.07 \ ms$ and $3.48\ mAh$ for CAMAL technique which are lowest over all the contemporary VPR frameworks. 
    $128 \times 384$ dimensional VLAD representation per image consumes $393KBytes$ memory.

    \renewcommand{\arraystretch}{1}
    \renewcommand{\tabcolsep}{1.8pt}    
    \begin{table}[t]
    	\centering
    	\caption{Evaluation metrics of the VPR aproaches.}
    	\label{table:vprComparison}
    	
    	
    	\label{table:comparison}
\begin{tabular}{|c|c|c|c|c|} 
	\hline
	\begin{tabular}[c]{@{}c@{}} \textbf{VPR }\\\textbf{Techniques}\\\textbf{$(Q=1622)$}\\\textbf{$(R=1622)$}\\ \end{tabular} & \begin{tabular}[c]{@{}c@{}}\textbf{Feature}\\\textbf{ Encoding }\\\textbf{Time}\\\textbf{$t_e$}\\\textbf{ (millisec)} \end{tabular} & \begin{tabular}[c]{@{}c@{}}\textbf{Feature }\\\textbf{ Matching }\\\textbf{Time}\\\textbf{$M_m$}\\\textbf{(millisec)} \end{tabular} & \begin{tabular}[c]{@{}c@{}}\textbf{Retrieval}\\\textbf{Time per }\\\textbf{query}\\\textbf{$M_q$}\\\textbf{(millisec)} \end{tabular} & \begin{tabular}[c]{@{}c@{}}\textbf{Power}\\\textbf{Consumption} \\\textbf{$U_m$ \textbar{} $U_e$ \textbar{} $t_m$ (millisec)}\\\textbf{$mAh$}\\\textbf{(miliampere-hours) } \end{tabular}  \\ 
	\hline
	\multicolumn{5}{|c|}{Intel Xeon(R) Gold 6134 CPU @ 3.20GHz with 32 cores, 64GB RAM}                                                                                                                                                                                                                                                                                                                                                                                                                                                                                                                                                                                                                                                      \\ 
	\hline
	\textit{NetVLAD}                                                                                                         & 770                                                                                                                                 & \textbf{0.005}                                                                                                                      & 778.11                                                                                                                               & \begin{tabular}[c]{@{}c@{}}0.036 \textbar{} 0.656 \textbar{} 8.11\\91.086 \end{tabular}                                                                                                    \\ 
	\hline
	\textit{Region-VLAD}                                                                                                     & 412                                                                                                                                 & 0.07                                                                                                                                & 525.54                                                                                                                               & \begin{tabular}[c]{@{}c@{}}0.031 \textbar{} 0.25 \textbar{} 113.54\\19.197 \end{tabular}                                                                                                   \\ 
	\hline
	\begin{tabular}[c]{@{}c@{}} \textit{Cross-Region}\\\textit{-BoW} \end{tabular}                                           & 830                                                                                                                                 & 160                                                                                                                                 & 260.35e3                                                                                                                             & \begin{tabular}[c]{@{}c@{}}0.1 \textbar{} 0.32 \textbar{} 259.52e3\\4724.99 \end{tabular}                                                                                                  \\ 
	\hline
	\textit{RMAC}                                                                                                            & 478                                                                                                                                 & 0.04                                                                                                                                & 542.88                                                                                                                               & \begin{tabular}[c]{@{}c@{}}0.371 \textbar{} 0.5 \textbar{} 64.88\\47.41 \end{tabular}                                                                                                      \\ 
	\hline
	\begin{tabular}[c]{@{}c@{}} Context \\Flexible \\Attention \end{tabular}                                                 & \textbf{90 }                                                                                                                        & 0.63                                                                                                                                & 1111.86                                                                                                                              & \begin{tabular}[c]{@{}c@{}}0.031 \textbar{} 0.5 \textbar{} 1021.86\\13.819 \end{tabular}                                                                                                   \\ 
	\hline
	\textit{CAMAL}                                                                                                           & 126.53                                                                                                                              & 0.07                                                                                                                                & \textbf{240.07 }                                                                                                                     & \begin{tabular}[c]{@{}c@{}}0.031 \textbar{} 0.125 \textbar{} 113.54\\\textbf{3.48 } \end{tabular}                                                                                          \\
	\hline
\end{tabular}
    	\vspace{-6mm}
    \end{table}
    
    In Table \ref{table:vprComparison}, power consumption and time required for feature encoding, matching and image retrieval for other VPR techniques are more than CAMAL excluding NetVLAD (low feature matching time) and Context Flexible Attention (low feature encoding time). Although, both NetVLAD and Context Flexible Attention employed deep VGG-16 but NetVLAD is implemented in tensorflow \cite{zaffar2019levelling} and returns a $1\times4096$ dimensional feature descriptor ($M_f + M_e + M_v = t_e = 770 \ ms$) taking $778.11 \ ms$ retrieval time and significant power use of $91.086 \ mAh$. Context Flexible Attention is implemented in Pytorch and $512 \times 14 \times 14$ dimensional feature ($M_f + M_e = 90 \ ms, \ M_v=0)$ consumes $401KBytes$ memory. The 3D feature is flattened to perform cosine distance matching (scipy pythonic library) that takes $M_m=0.63 \ ms$. It is evident that excluding power use, Region-VLAD and RMAC have comparable feature encoding, matching and retrieval times but Cross-Region-BoW is the most resource hungry due to intense $M_m$ feature matching time (see \cite{zaffar2019levelling}).
    
    The logarithmic plots of $M_q$ retrieval time and $mAh$ power consumption against $R$ reference images in Fig. \ref{figure:imgRet} \& Table \ref{table:vprComparison} indicate that although Context Flexible Attention has the lowest feature encoding time but its not optimal for battery-powered large-scale VPR applications; at large database size, the technique takes too much retrieval/recognition time, i.e., approximately $1.11 \ s$ retrieval time per query for $R=1622$ ($4.6x$ more than CAMAL) and power consumption of $13.819 \ mAh$ for $Q=1622$ ($3.9x$ more than CAMAL). 

 
    \subsubsection{\textbf{Precision-Recall Characteristics}}    
\begin{figure}[t]
	\centering
	\vspace{-2mm}
	\includegraphics[width=1\linewidth, height=1\linewidth,keepaspectratio]{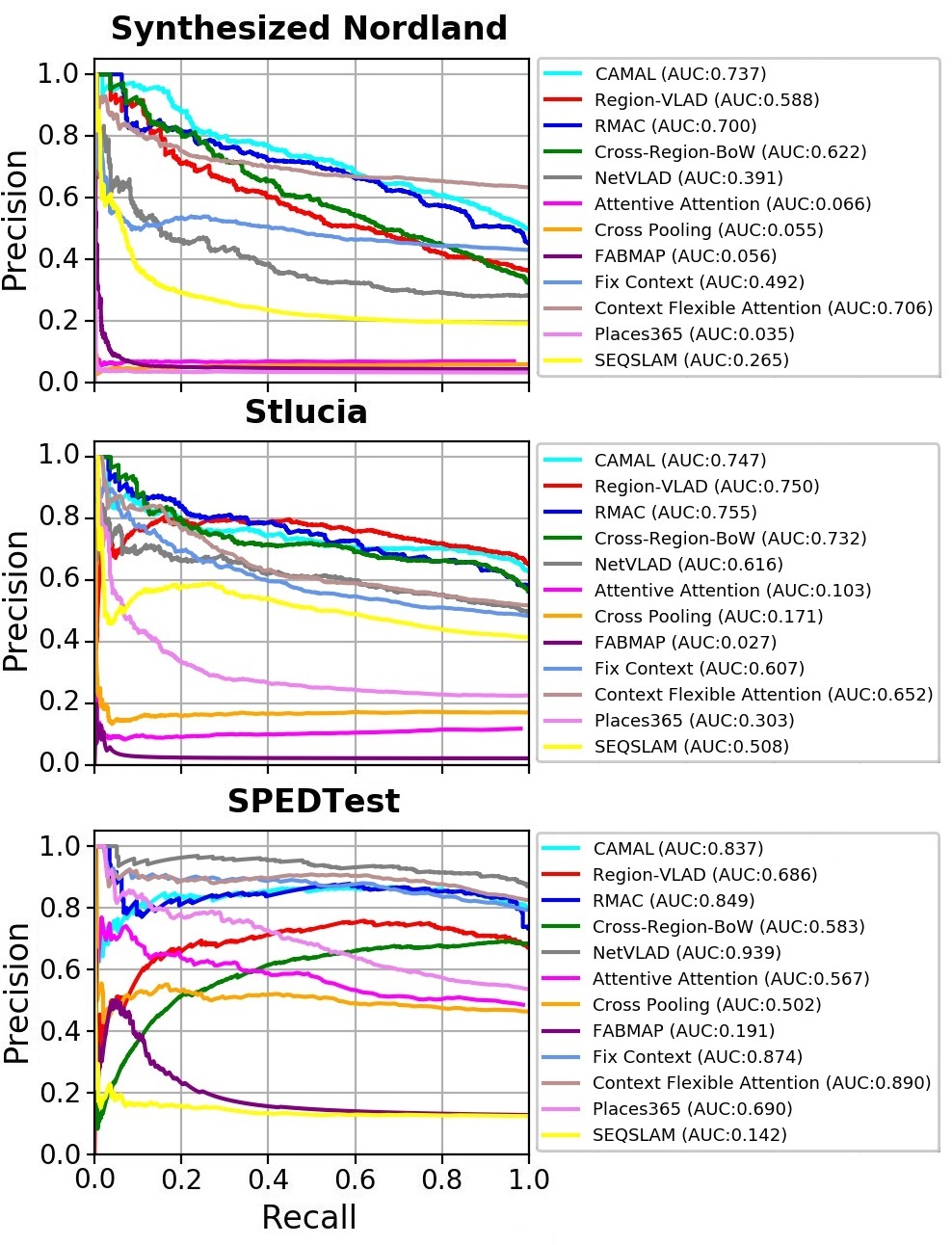}
	\caption{AUC-PR curves of the benchmark datasets evaluated on various VPR algorithms are shown here. Our CAMAL framework exhibits better \& comparable performance against SOTA Context Flexible Attention \cite{chen2018learning}.
	}
	\label{figure:allDatasets}
	\vspace{-6mm}
	
\end{figure}

    More area the PR-curve covers, better the recognition performance from the technique can be expected. Fig. \ref{figure:allDatasets} reports the AUC-PR performance for the benchmark datasets evaluated on the VPR techniques. It is visible that our proposed approach have shown better and comparable AUC-PR performance against SOTA VPR algorithms. 
    A closer look at CAMAL results indicate that the captured place-specific convolutional attentions boost up the overall recognition performance. It suggests that the employment of multiple convolutional feature are efficient in dealing with environment invariant-VPR used for global localization of mobile/aerial robotics.

    Other VPR frameworks including Region-VLAD, Cross-Region-BoW and RMAC for St.Lucia dataset have shown similar performance as CAMAL. However, their performance degrades for SPEDTest and Synthesized Nordland datasets which experience seasonal change. It is observed that under moderate conditional variation, region-based frameworks achieve par recognition performance. NetVLAD showcases nearly the similar PR-characteristic as Fixed Context and Context Flexible Attention.    
    However, despite the SOTA performance on SPEDTest, NetVLAD underperformed on Synthesized Nordland dataset with a big margin. 
    One reason could be its sensitivity towards the existence of perceptual aliasing (homogeneous scenes) which points out that the CNN's training dataset lacked in condition variance. For diverse SPEDTest, each test image has three matched reference images only thus recognized the places efficiently.

     Cross-Region-BoW and Region-VLAD have shown an average performance on these datasets. It is observed that due to ImageNet-centric nature of VGG-16, the cross-regional approach concentrates more on objects whereas scene-centric nature of AlexNet365 integrated with Region-VLAD might have forced it to consider dynamic instances e.g. sky as a valuable region for distinguishing the scenes which leads to incorrect match for some places. 
    However, under moderate condition and viewpoint variant \textit{St.Lucia} dataset, it exhibits better performance.
    It is visible that the worst performance of FABMAP is consistent for all the datasets. It is because FABMAP used viewpoint-invariant SURF feature detector which is sensitive under conditional variation. It is interesting that SEQSLAM with its better appearance tackling and whole image-based matching approach shown inferior performance under SPEDTest. It is probably because the places exhibit diverse environment, three reference per test image and sequence-based matching requirement is violated. 
    
    Despite the better performances of Cross-Pool and Attention Attentive approach in other vision-based tasks, both under-performed in St.Lucia and Synthesized Nordland datasets. This highlights the difference in other image retrieval/classification systems from specific place recognition. 
    However, their better performance on SPEDTest point towards the importance of CNN training since authors fine-tuned the Attention Attentive on SPED. Better and comparable performance of CAMAL on all the dataset highlights the usefulness and generalization power of lightweight contextual attentions over D-CNNs based VPR algorithms \cite{chen2018learning}.

    \vspace{-1mm}
    \subsection{Quantitative Analysis}
    A deep analysis suggests that although HybridNet \cite{chen2017deep} and Context Flexible Attention \cite{chen2018learning} CNN models are fine-tuned on SPED but training parameters such as learning rates are kept different; dual learning rates approach was employed in \cite{chen2018learning}. The weight decays and iterations also kept different from the values set for HyridNet and SPED-centric VGG-16. Employing three convolutional layers, deep multi-layer features of \cite{chen2018learning} might have learned more powerful condition-invariant features given different trainable configuration and hence, exhibits better performance for SPEDTest. However, under seasonal changes coupled with perceptual aliasing (Synthesized Nordland), the performance degrades. It should be noted that our proposed CAMAL approach employed only two convolutional layers and still delivers a comparable performance across all the datasets in terms of AUC-PR curves with $4x$ reduction in resource \& power utilization.
    
\begin{figure}[t]
		\centering
	    \includegraphics[width=1\linewidth, height=1\linewidth,keepaspectratio]{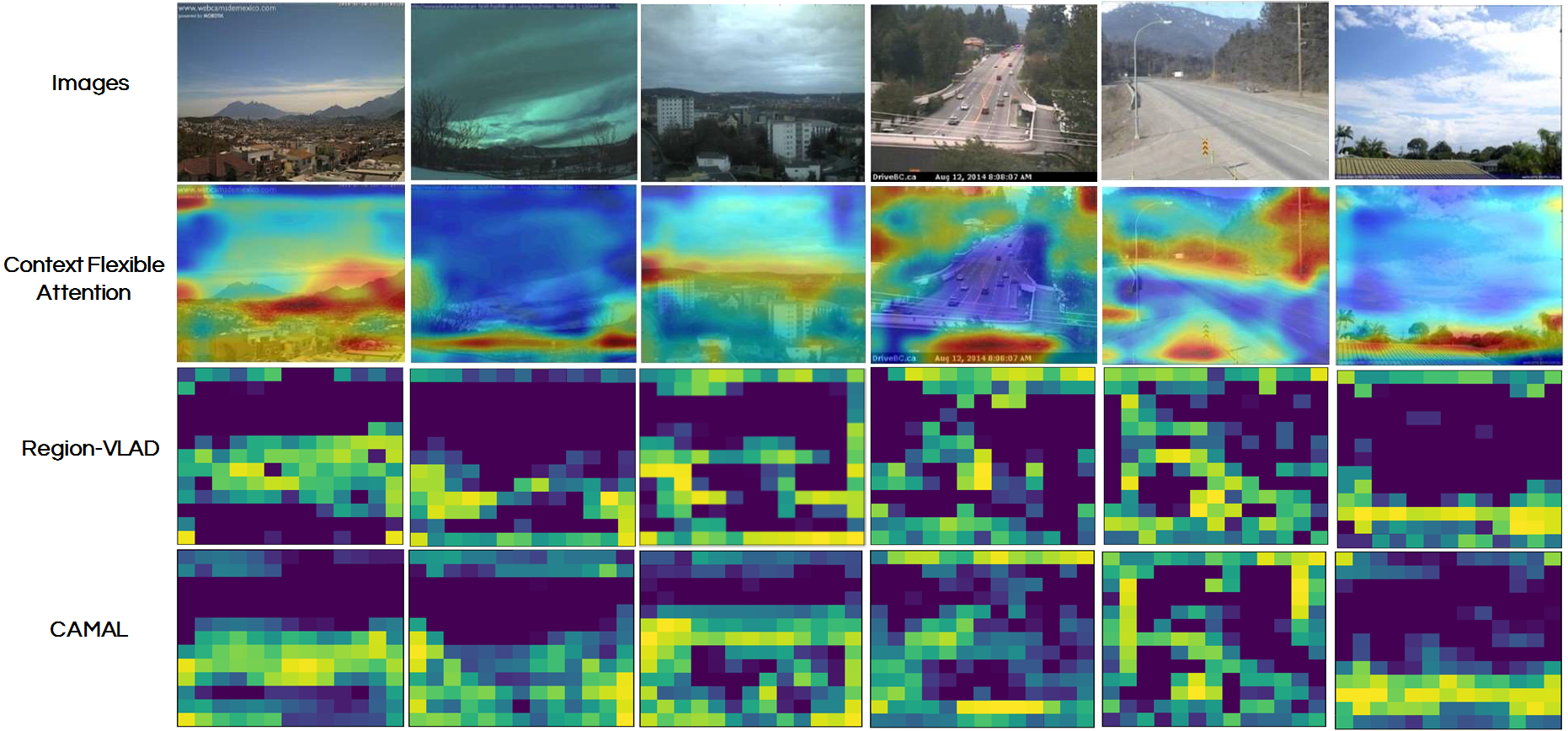}
		\caption{Sample place-specific regional attentions identified by our proposed computation- \& energy-efficient CAMAL approach against the contemporary VPR algorithms are illustrated here. 
		} 
		\label{figure:m_regVladVsContxtFlex}
		\vspace{-5mm}
		
	\end{figure}

    Furthermore, Fig. \ref{figure:m_regVladVsContxtFlex} analyzes and differentiates the convolutional attentions captured by CAMAL from other SOTA Context Flexible Attention \cite{chen2018learning} and Region-VLAD \cite{khaliqholistic}. Both Context Flexible Attention and CAMAL capture place-specific structures such as houses, street lights while filtering out confusing areas including clouds, vehicles etc. It can be seen that Region-VLAD sometimes include sky and other dynamic instances as vital regions. Our CAMAL VPR technique captures place-specific regional attentions from a shallow CNN place recognition-centric HybridNet robust under strong condition and mild viewpoint variations. 
    
    Fixed Context, NetVLAD and Places365 techniques exhibit better AUC-PR performance for SPEDTest and St.Lucia datasets. It implies that they are sensitive towards perceptual aliasing experienced in Synthesized Nordland dataset. Region-VLAD and Places365 techniques are integrated on scene-centric CNN model but there is a AUC-PR performance difference for Synthesized Nordland and St.Lucia datasets. Further investigations suggest that the performance improvement cannot be achieved by directly employing the pretrained task-dependent CNNs rather it's the feature pooling approach that also plays a vital role in improving the recognition performance, i.e., by focusing on place-specific convolutional activations which is one of the motivation of this work and other motivation focused on lower computation and energy consumption needed for battery-operated robotic use-cases. Datasets and AUC-PR results are placed at \cite{ahmedest61VPR}. 

\vspace{-2mm}
    \section{Conclusion}
    Despite the recent SOTA performance of D-CNNs for VPR, 
    the high computation and power consumption limit their practical deployment for battery-operated mobile and aerial robotics. Achieving superior performance with lightweight CNN architectures is thus desirable, but a challenging problem. In this paper, a multi-layer place-specific attention approach is presented that combines salient regions from multiple convolutional layers of CNN SPED-centric HybridNet. The proposed CAMAL framework captures persistent regional attentions robust under large scale environment invariant-VPR. 
    At $4x$ power reduction, evaluation on several challenging datasets confirm the dominance over contemporary VPR algorithms in terms of AUC-PR curves with $4x$ quicker retrieval/recognition time performance.
    
    
    In future, we will incorporate the proposed multi-layer attention block in a shallow feed forward neural network and fine-tune the CNN model on a large-scale place recognition dataset employing object recognition or semantic segmentation and then learn information about the staticness of those objects or semantic regions. It should learn image regions invariant to strong viewpoint and condition variations with reduction in the feature encoding and image retrieval times.

    \vspace{-2mm}
    \section*{ACKNOWLEDGMENT}

    We gratefully acknowledge the support of Dr. Zetao Chen (Facebook Zurich, Switzerland) for providing details of \cite{chen2018learning}. 
    \vspace{-3mm}
    
    
    \bibliographystyle{IEEEtran}
	\bibliography{root}

\begin{thebibliography}{10}
\providecommand{\url}[1]{#1}
\csname url@samestyle\endcsname
\providecommand{\newblock}{\relax}
\providecommand{\bibinfo}[2]{#2}
\providecommand{\BIBentrySTDinterwordspacing}{\spaceskip=0pt\relax}
\providecommand{\BIBentryALTinterwordstretchfactor}{4}
\providecommand{\BIBentryALTinterwordspacing}{\spaceskip=\fontdimen2\font plus
\BIBentryALTinterwordstretchfactor\fontdimen3\font minus
  \fontdimen4\font\relax}
\providecommand{\BIBforeignlanguage}[2]{{%
\expandafter\ifx\csname l@#1\endcsname\relax
\typeout{** WARNING: IEEEtran.bst: No hyphenation pattern has been}%
\typeout{** loaded for the language `#1'. Using the pattern for}%
\typeout{** the default language instead.}%
\else
\language=\csname l@#1\endcsname
\fi
#2}}
\providecommand{\BIBdecl}{\relax}
\BIBdecl

\bibitem{lowry2016visual}
S.~Lowry, N.~S{\"u}nderhauf \emph{et~al.}, ``Visual place recognition: A
  survey,'' \emph{IEEE Transactions on Robotics}, vol.~32, no.~1, pp. 1--19,
  2016.

\bibitem{cadena2016past}
C.~Cadena, L.~Carlone \emph{et~al.}, ``Past, present, and future of
  simultaneous localization and mapping: Toward the robust-perception age,''
  \emph{IEEE Transactions on Robotics}, vol.~32, no.~6, pp. 1309--1332, 2016.

\bibitem{bay2006surf}
H.~Bay, T.~Tuytelaars \emph{et~al.}, ``Surf: Speeded up robust features,'' in
  \emph{Proc. European Conference on Computer Vision}, 2006, pp. 404--417.

\bibitem{ke2004pca}
Y.~Ke, R.~Sukthankar \emph{et~al.}, ``Pca-sift: A more distinctive
  representation for local image descriptors,'' \emph{CVPR (2)}, vol.~4, pp.
  506--513, 2004.

\bibitem{dalal2005histograms}
N.~Dalal \emph{et~al.}, ``Histograms of oriented gradients for human
  detection,'' in \emph{CVPR}, vol.~1.\hskip 1em plus 0.5em minus 0.4em\relax
  IEEE Computer Society, 2005, pp. 886--893.

\bibitem{chen2014convolutional}
Z.~Chen, O.~Lam, A.~Jacobson, and M.~Milford, ``Convolutional neural
  network-based place recognition,'' \emph{Australasian Conference on Robotics
  and Automation}, 2014.

\bibitem{sunderhauf2015performance}
N.~S{\"u}nderhauf, S.~Shirazi, F.~Dayoub \emph{et~al.}, ``On the performance of
  convnet features for place recognition,'' in \emph{IEEE International
  Conference on Intelligent Robots and Systems}, 2015, pp. 4297--4304.

\bibitem{sunderhauf2015place}
N.~S{\"u}nderhauf, S.~Shirazi, A.~Jacobson, F.~Dayoub, E.~Pepperell, B.~Upcroft
  \emph{et~al.}, ``Place recognition with convnet landmarks: Viewpoint-robust,
  condition-robust, training-free,'' \emph{Proc. Robotics: Science and Systems
  Conference}, 2015.

\bibitem{wang2017residual}
F.~Wang, M.~Jiang, C.~Qian, S.~Yang, C.~Li, H.~Zhang \emph{et~al.}, ``Residual
  attention network for image classification,'' in \emph{Proc. IEEE Conference
  on Computer Vision and Pattern Recognition}, 2017, pp. 3156--3164.

\bibitem{wang2017multi}
P.~Wang, L.~Liu, C.~Shen, Z.~Huang \emph{et~al.}, ``Multi-attention network for
  one shot learning,'' in \emph{Proc. IEEE Conference on Computer Vision and
  Pattern Recognition}, 2017, pp. 2721--2729.

\bibitem{krizhevsky2012imagenet}
A.~Krizhevsky, I.~Sutskever, and G.~E. Hinton, ``Imagenet classification with
  deep convolutional neural networks,'' in \emph{Annual Conference in Neural
  Information Processing Systems}, 2012, pp. 1097--1105.

\bibitem{zhou2017places}
B.~Zhou, A.~Lapedriza \emph{et~al.}, ``Places: A 10 million image database for
  scene recognition,'' \emph{IEEE Transactions on Pattern Analysis and Machine
  Intelligence}, 2017.

\bibitem{torii201524}
A.~Torii, R.~Arandjelovic, J.~Sivic, M.~Okutomi, and T.~Pajdla, ``24/7 place
  recognition by view synthesis,'' in \emph{IEEE Conference on Computer Vision
  and Pattern Recognition}, 2015, pp. 1808--1817.

\bibitem{chen2017only}
Z.~Chen \emph{et~al.}, ``Only look once, mining distinctive landmarks from
  convnet for visual place recognition,'' in \emph{IEEE International
  Conference on Intelligent Robots and Systems}, 2017, pp. 9--16.

\bibitem{simonyan2014very}
K.~Simonyan \emph{et~al.}, ``Very deep convolutional networks for large-scale
  image recognition,'' \emph{International Conference on Learning
  Representations}, 2015.

\bibitem{chen2018learning}
Z.~Chen, L.~Liu, I.~Sa, Z.~Ge, and M.~Chli, ``Learning context flexible
  attention model for long-term visual place recognition,'' \emph{IEEE Robotics
  and Automation Letters}, vol.~3, no.~4, pp. 4015--4022, 2018.

\bibitem{chen2017deep}
Z.~Chen, A.~Jacobson, N.~S{\"u}nderhauf, B.~Upcroft, L.~Liu, C.~Shen, I.~Reid,
  and M.~Milford, ``Deep learning features at scale for visual place
  recognition,'' in \emph{IEEE International Conference on Robotics and
  Automation}, 2017, pp. 3223--3230.

\bibitem{khaliqholistic}
A.~{Khaliq}, S.~{Ehsan}, Z.~{Chen}, M.~{Milford} \emph{et~al.}, ``A holistic
  visual place recognition approach using lightweight cnns for significant
  viewpoint and appearance changes,'' \emph{IEEE Transactions on Robotics}, pp.
  1--9, 2019.

\bibitem{jegou2010aggregating}
H.~J{\'e}gou, M.~Douze, C.~Schmid, and P.~P{\'e}rez, ``Aggregating local
  descriptors into a compact image representation,'' in \emph{IEEE Conference
  on Computer Vision and Pattern Recognition}, 2010, pp. 3304--3311.

\bibitem{cummins2008fab}
M.~Cummins and P.~Newman, ``{FAB-MAP}: Probabilistic localization and mapping
  in the space of appearance,'' \emph{International Journal of Robotics
  Research}, vol.~27, no.~6, pp. 647--665, 2008.

\bibitem{milford2012seqslam}
M.~J. Milford \emph{et~al.}, ``{SeqSLAM}: Visual route-based navigation for
  sunny summer days and stormy winter nights,'' in \emph{IEEE International
  Conference on Robotics and Automation}, 2012, pp. 1643--1649.

\bibitem{tolias2015particular}
G.~Tolias, R.~Sicre, and H.~J{\'e}gou, ``Particular object retrieval with
  integral max-pooling of cnn activations,'' \emph{Proc. International
  Conference on Learning Representations}, 2016.

\bibitem{babenko2015aggregating}
A.~Babenko and V.~Lempitsky, ``Aggregating local deep features for image
  retrieval,'' in \emph{IEEE International Conference on Computer Vision},
  2015, pp. 1269--1277.

\bibitem{liu2017cross}
L.~Liu, C.~Shen, and A.~van~den Hengel, ``Cross-convolutional-layer pooling for
  image recognition,'' \emph{IEEE Transactions on Pattern Analysis and Machine
  Intelligence}, vol.~39, no.~11, pp. 2305--2313, 2017.

\bibitem{chu2017multi}
X.~Chu, W.~Yang, W.~Ouyang, C.~Ma \emph{et~al.}, ``Multi-context attention for
  human pose estimation,'' in \emph{Proc. IEEE Conference on Computer Vision
  and Pattern Recognition}, 2017, pp. 1831--1840.

\bibitem{jin2017learned}
H.~J. Kim, E.~Dunn \emph{et~al.}, ``Learned contextual feature reweighting for
  image geo-localization,'' in \emph{IEEE Conference on Computer Vision and
  Pattern Recognition}, 2017, pp. 2136--2145.

\bibitem{rosten2006machine}
E.~Rosten and T.~Drummond, ``Machine learning for high-speed corner
  detection,'' in \emph{ECCV}.\hskip 1em plus 0.5em minus 0.4em\relax Springer,
  2006, pp. 430--443.

\bibitem{oliva2006building}
A.~Oliva and A.~Torralba, ``Building the gist of a scene: The role of global
  image features in recognition,'' \emph{Progress in brain research}, vol. 155,
  pp. 23--36, 2006.

\bibitem{sivic2003video}
J.~Sivic \emph{et~al.}, ``Video google: A text retrieval approach to object
  matching in videos,'' in \emph{IEEE International Conference on Computer
  Vision}, 2003, p. 1470.

\bibitem{panphattarasap2016visual}
P.~Panphattarasap and A.~Calway, ``Visual place recognition using landmark
  distribution descriptors,'' in \emph{Asian Conference on Computer
  Vision}.\hskip 1em plus 0.5em minus 0.4em\relax Springer, 2016, pp. 487--502.

\bibitem{arandjelovic2016netvlad}
R.~Arandjelovic \emph{et~al.}, ``Net{VLAD}: {CNN} architecture for weakly
  supervised place recognition,'' in \emph{IEEE Conference on Computer Vision
  and Pattern Recognition}, 2016, pp. 5297--5307.

\bibitem{zaffar2018memorablemaps}
M.~Zaffar, S.~Ehsan, M.~Milford, and K.~McDonald-Maier, ``Memorable maps: A
  framework for re-defining places in visual place recognition,'' \emph{\tt
  arXiv:1811.03529 [cs.CV]}, 2018.

\bibitem{facil2019condition}
J.~M. Facil, D.~Olid \emph{et~al.}, ``Condition-invariant multi-view place
  recognition,'' \emph{\tt arXiv:1902.09516 [cs.CV]}, 2019.

\bibitem{8972582}
M.~{Zaffar}, S.~{Ehsan}, M.~J. {Milford}, and K.~{McDonald-Maier}, ``Cohog: A
  light-weight, compute-efficient and training-free visual place recognition
  technique for changing environments,'' \emph{IEEE Robotics and Automation
  Letters}, pp. 1--1, 2020.

\bibitem{siagian2009biologically}
C.~Siagian and L.~Itti, ``Biologically inspired mobile robot vision
  localization,'' \emph{IEEE Transactions on Robotics}, vol.~25, no.~4, pp.
  861--873, 2009.

\bibitem{chen2016attention}
L.-C. Chen, Y.~Yang, J.~Wang, W.~Xu \emph{et~al.}, ``Attention to scale:
  Scale-aware semantic image segmentation,'' in \emph{Proc. IEEE conference on
  Computer Vision and Pattern Recognition}, 2016, pp. 3640--3649.

\bibitem{kim2017learned}
H.~J. Kim, E.~Dunn \emph{et~al.}, ``Learned contextual feature reweighting for
  image geo-localization,'' in \emph{IEEE Conference on Computer Vision and
  Pattern Recognition}.\hskip 1em plus 0.5em minus 0.4em\relax IEEE, 2017, pp.
  3251--3260.

\bibitem{Neubert2014SuperpixelbasedAC}
P.~Neubert, N.~S{\"u}nderhauf, and P.~Protzel, ``Superpixel-based appearance
  change prediction for long-term navigation across seasons,'' \emph{Robotics
  and Autonomous Systems}, vol.~69, pp. 15--27, 2014.

\bibitem{yu2017multi}
D.~Yu, J.~Fu, T.~Mei, and Y.~Rui, ``Multi-level attention networks for visual
  question answering,'' in \emph{IEEE Conference on Computer Vision and Pattern
  Recognition}, 2017, pp. 4187--4195.

\bibitem{cummins2009highly}
M.~Cummins, ``Highly scalable appearance-only slam-fab-map 2.0,'' \emph{Proc.
  Robotics: Sciences and Systems (RSS), 2009}, 2009.

\bibitem{zaffar2019levelling}
M.~Zaffar, A.~Khaliq, S.~Ehsan \emph{et~al.}, ``Levelling the playing field: A
  comprehensive comparison of visual place recognition approaches under
  changing conditions,'' \emph{arXiv preprint arXiv:1903.09107}, 2019.

\bibitem{zaffar2019state}
M.~Zaffar, A.~Khaliq, S.~Ehsan, M.~Milford, K.~Alexis \emph{et~al.}, ``Are
  state-of-the-art visual place recognition techniques any good for aerial
  robotics?'' \emph{arXiv preprint arXiv:1904.07967}, 2019.

\bibitem{hanley1982meaning}
J.~A. Hanley and B.~J. McNeil, ``The meaning and use of the area under a
  receiver operating characteristic {(ROC)} curve.'' \emph{Radiology}, vol.
  143, no.~1, pp. 29--36, 1982.

\bibitem{ahmedest61VPR}
``Results and datasets,''
  \url{https://github.com/Ahmedest61/CNN-Region-VLAD-VPR/}.

\end{thebibliography}

\end{document}